\documentclass[10pt,twocolumn,letterpaper]{article}

\usepackage[pagenumbers]{cvpr} 

\usepackage{times}
\usepackage{epsfig}
\usepackage{graphicx}
\usepackage{amsmath}
\usepackage{amssymb}
\usepackage{booktabs}
\usepackage{multirow}
\usepackage{siunitx}
\usepackage{bm}
\usepackage{makecell}
\usepackage{threeparttable}
\usepackage{tablefootnote}

\usepackage[pagebackref,breaklinks,colorlinks]{hyperref}

\newcommand{\textBF}[1]{%
    \pdfliteral direct {2 Tr 0.3 w} 
     #1%
    \pdfliteral direct {0 Tr 0 w}%
}

\usepackage{capt-of,etoolbox}
\usepackage{enumitem}

\usepackage[capitalize]{cleveref}
\crefname{section}{Sec.}{Secs.}
\Crefname{section}{Section}{Sections}
\Crefname{table}{Table}{Tables}
\crefname{table}{Tab.}{Tabs.}

\def\cvprPaperID{20} 
\def\confName{CVPR}
\def\confYear{2023}

\begin{document}

\title{Unsupervised Style-based Explicit 3D Face Reconstruction from Single Image}


\author{Heng Yu \quad Zolt\'{a}n \'{A}. Milacski \quad L\'{a}szl\'{o} A. Jeni  \vspace{4pt}\\
	Robotics Institute, Carnegie Mellon University \\
    {\tt\small \{hengyu, zmilacsk\}@andrew.cmu.edu} \quad {\tt\small laszlojeni@cmu.edu} \\
}

\twocolumn[{%
\renewcommand\twocolumn[1][]{#1}%
\maketitle
\begin{center}
    \centering
    \includegraphics[width=0.81\textwidth]{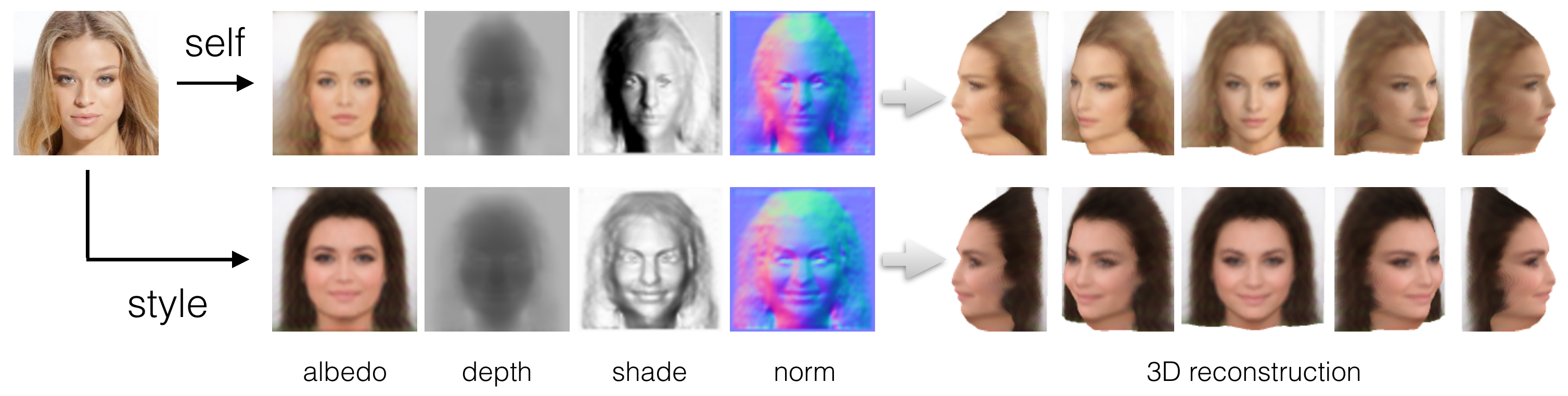}
    \captionof{figure}{Given a single 2D image, we reconstruct the 3D object (top) by predicting the features of the rendering process: albedo, depth, shading and surface normals. Then the method can synthesize images from an arbitrary viewpoint with a new style, potentially including changes in both shape and appearance (bottom).}
    \label{fig:overview}
\end{center}%
}]


\begin{abstract}
   Inferring 3D object structures from a single image is an ill-posed task due to depth ambiguity and occlusion.
   Typical resolutions in the literature include leveraging 2D or 3D ground truth for supervised learning, as well as imposing hand-crafted symmetry priors or using an implicit representation to hallucinate novel viewpoints for unsupervised methods.
   In this work, we propose a general adversarial learning framework for solving Unsupervised 2D to Explicit 3D Style Transfer (UE3DST).
   Specifically, we merge two architectures: the unsupervised explicit 3D reconstruction network of Wu et al.\ and the Generative Adversarial Network (GAN) named StarGAN-v2.
   We experiment across three facial datasets (Basel Face Model, 3DFAW and CelebA-HQ) and show that our solution is able to outperform well established solutions such as DepthNet in 3D reconstruction and Pix2NeRF in conditional style transfer, while we also justify the individual contributions of our model components via ablation.
   In contrast to the aforementioned baselines, our scheme produces features for explicit 3D rendering, which can be manipulated and utilized in downstream tasks.
\end{abstract}

\section{Introduction}
\label{sec:intro}

Reconstructing the underlying 3D structure of objects from few 2D images or even a single 2D image is a promising research area, since we live in the 3D world.
Inferring accurate 3D models of human faces, bodies and other objects can greatly promote the development of several fields like animation, video games, virtual reality, augmented reality and the metaverse.
This task is difficult, however, due to various forms of underdetermination, including depth ambiguity and occlusion.
The straighforward resolution to these problems is applying supervised learning with 2D/3D ground truth information,
keypoints \cite{kanazawa2018learning,novotny2022keytr} or depth maps \cite{jung2009novel, wang2019pseudo}.
However, it is hard and sometimes even impossible to collect such supervisory signals, as this requires considerable human effort.
Hence, recent techniques \cite{zhang1999shape, mukherjee1995shape, thrun2005shape}  exploit unsupervised learning to leverage human prior knowledge about the natural objects in the world.
This alleviates the need for annotated data, but the prior has to be chosen wisely for each object class.
E.g., human faces are roughly symmetric, and exploiting this as a prior yields considerable improvements \cite{wu2020unsupervised}.

Besides reconstructing the 3D structure, one may be also interested in altering its style to one that is different from that of the original input image.
Consider, e.g., supporting {}---{} or even replacing {}---{} CG artists by automatically changing a 3D human model's gender, haircut, skin color or age.
Interestingly, \emph{Unsupervised 2D to Explicit 3D Style Transfer (UE3DST)} is a less discussed problem in the literature.
By \emph{explicit}, we mean that the method extracts a set of features, which allow recovering the 3D shape via using an explicit rendering function \cite{wu2020unsupervised}.
Unfortunately, existing methods are either (i) implicit 3D, or (ii) supervised.
The widely adopted solution for (i) is hallucinating novel 3D viewpoints using some form of a Generative Adversarial Network (GAN) \cite{goodfellow2014generative} or a Neural Radiance Field (NeRF) \cite{mildenhall2020nerf}.
An approach for (ii) is SofGAN \cite{chen2022sofgan}, which recovers explicit 3D and can change styles such as pose, shape, and texture; however, it requires semantic segmentation maps as ground truth.
While these techniques for (i)-(ii) are able to produce remarkable results on their own benchmarks, they are less applicable due to (i) lacking explicit 3D features and (ii) relying entirely on supervision.

In this paper, we propose a framework to tackle the UE3DST problem by unifying the solutions of (a) unsupervised 3D reconstruction and (b) style transfer.
Specifically, we introduce an unsupervised end-to-end trainable framework (Fig.~\ref{fig:overview}) that utilizes a GAN  to generate an explicit 3D representation given a single input image and a style.
While our framework supports arbitrary choices of architectures, throughout our quantitative experiments, we integrate the methods of Wu et al.\ \cite{wu2020unsupervised} and StarGAN-v2 \cite{choi2020stargan}.
We test the hypothesis whether our combined method achieves better performance compared to existing methods, including DepthNet \cite{moniz2018unsupervised} for (a) and Pix2NeRF \cite{cai2022pix2nerf} for (b), as our solution can compete in their respective benchmarks.
We perform our quantitative and qualitative experiments on three facial datasets: CelebA-HQ \cite{karras2017progressive}, 3DFAW \cite{gross2010multi, jeni2015dense, zhang2013high, zhang2014bp4d} and Basel Face Model (BFM) \cite{paysan20093d}.
Lastly, we also perform an ablation study to test the efficiencies of different components of our model.
The source code enabling the reproduction of our results will be made publicly available.

\section{Related Works}
In this section, we review 2D to 3D reconstruction and style transfer methods.
Let us first introduce a categorizations of the former methods.


\emph{Explicit} techniques extract features of the 3D shape, which facilitate the recovery of the shape through an explicit rendering function.
These features can then be post-processed or used for downstream tasks.
In contrast, \emph{implicit} approaches do not yield an explicit 3D representation of the object, but are still capable of producing an implicit one that cannot be manipulated and used easily.


\subsection{2D to 3D Reconstruction}
\textbf{Shallow Explicit 3D Reconstruction}
Traditional methods such as Structure from Motion (SfM) \cite{ullman1979interpretation} use geometric relationships between matching 2D keypoints in multiple views of the same rigid scene to recover the 3D structure.
Non-Rigid SfM \cite{bregler2000recovering, sidhu2020neural} extends this to deformable objects, but it still requires 2D key point annotations for training and testing.
In addition, other information such as shading \cite{zhang1999shape} and symmetry \cite{mukherjee1995shape, thrun2005shape} have been proven to be useful for 3D recovery.

\textbf{Deep Explicit 3D Reconstruction}
With the recent development of deep learning technology, superior 3D reconstruction methods have emerged, facilitating the learning of a suitable shape prior from sufficient training data in order to alleviate the ill-posedness of 3D recovery \cite{ranjan2018generating, wu2016learning}.
\emph{Supervised} methods map images to their paired target variables, which include: 3D ground truth (e.g., meshes \cite{wang2018pixel2mesh}, voxels \cite{shin2018pixels}, points clouds \cite{navaneet2020image}), 3D shape models, video \cite{agrawal2015learning, novotny2017learning, wang2018learning, zhou2017unsupervised, sun2021neuralrecon}, keypoints \cite{chen2019unsupervised, kanazawa2018learning, suwajanakorn2018discovery}, 2D silhouettes, stereo image pairs \cite{godard2017unsupervised} or even just object classes \cite{thewlis2018modelling, kanazawa2018learning}.
In contrast, \emph{unsupervised} schemes learn from raw images only, such as: 3D lifting for bottleneck codes of deformation field autoencoders \cite{sahasrabudhe2019lifting}, adversarial 3D mesh reconstruction \cite{szabo2019unsupervised}, or using a network with disentangled rendering features with symmetry regularization for face depth and albedo \cite{wu2020unsupervised}.


\textbf{Deep Implicit 3D Reconstruction}
As mentioned above, implicit schemes only output an implicit function representation of the object that is not useful on its own, requiring further processing, e.g., conversion to explicit form (see Marching Cubes \cite{lorensen1987marching}) or alternative ways of rendering (e.g., deep network based 3D viewpoint to 2D image hallucination, including differentiable ray casting). Here, we specifically focus on novel 3D viewpoint synthesis, which is inherently \emph{unsupervised}.
\\
\emph{Generative Adversarial Networks.}
A Generative Adversarial Network (GAN) \cite{goodfellow2014generative} consists of two competing components: a generator and a discriminator.
The generator takes a given latent (and optionally conditional) distribution as input, outputs `generated' examples, and tries make them indistinguishable from the `real' training set for the discriminator.
The two networks are trained in alternating manner, thus both are improving the other by making the task of the counterpart more and more difficult during training.
Note, however, that mode dropping should be avoided: the generator has to produce diverse examples instead of just fooling the discriminator by replicating few training samples.
GANs for viewpoint generation include
image conditional 3D-aware networks \cite{aizawa2021agnostic, kossaifi2018gagan, geng2018warp, nguyen2018rendernet, rhodin2018unsupervised, sitzmann2019deepvoxels} that accept viewpoint features as input and use them to transform the hidden representation in 3D.
3D aware images can be generated unconditionally from noise as well \cite{shi2021lifting, zhang2020image, deng2020disentangled, nguyen2019hologan}.
\\
\emph{Neural Radiance Fields.}
Neural Radiance Field (NeRF) \cite{mildenhall2020nerf} exploits differentiable volume rendering to learn density and color, and they must be trained on sufficiently large single-scene multi-view image sets, limiting applicability.
\\
\emph{Hybrid Architectures.}
Recent works incorporate a NeRF into their GAN architectures \cite{schwarz2020graf, niemeyer2021giraffe, gu2021stylenerf, deng2021gram, chan2021pi, cai2022pix2nerf}.
Typically, the generator is a NeRF and the discriminator classifies its hallucinated 2D images.
However, these procedures are restricted to scene-specific or category-specific implicit representations.

Our work differs from these schemes, since in contrast to recovering the original 3D structure, we also consider changing the 3D style.

\subsection{Style Transfer}

\textbf{2D Style Transfer}
GANs achieve remarkable performance on 2D image generation \cite{brock2018large, berthelot2017began, zhang2019self} and style transfer (sometimes also referred to as image-to-image translation) \cite{pix2pix2017, CycleGAN2017, karras2019style, choi2020stargan}. Pix2pix \cite{pix2pix2017} solves the problem using conditional adversarial networks. Cycle-GAN \cite{CycleGAN2017} uses a special cyclic architecture and uses cycle consistency losses to achieve remarkable performance. Karras et al. \cite{karras2019style} achieve unsupervised separation of high-level attributes and scale-specific control of the synthesis via Adaptive Instance Normalization (AdaIN) \cite{huang2017arbitrary, karras2019style} at each convolutional layer. Choi et al. \cite{choi2020stargan} proposed StarGAN-v2 consisting of four modules, which significantly improve the diversity of generated images under multiple domains.

\textbf{Explicit 3D Style Transfer}
SofGAN \cite{chen2022sofgan} achieves explicit 3D reconstruction and transferring styles such as pose, shape, and texture. However, it is supervised: it requires semantic segmentation maps as ground truth, limiting applicability.

\textbf{Implicit 3D Style Transfer}
HoloGAN \cite{nguyen2019hologan} learns to synthesize 3D-aware images from unlabeled 2D images and disentangles shape and appearance.
$\pi$-GAN \cite{chan2021pi} leverages a SIREN-based NeRF to get multi-view consistency and high quality images.
Pix2NeRF \cite{cai2022pix2nerf} is conditioned on a single input image based on $\pi$-GAN and disentangles pose and content.

The approaches listed here are not unsupervised and explicit at the same time, whereas our proposed method can tackle this scenario.

\section{Methods}

\begin{figure*}
\centering
\includegraphics[width=0.6\textwidth]{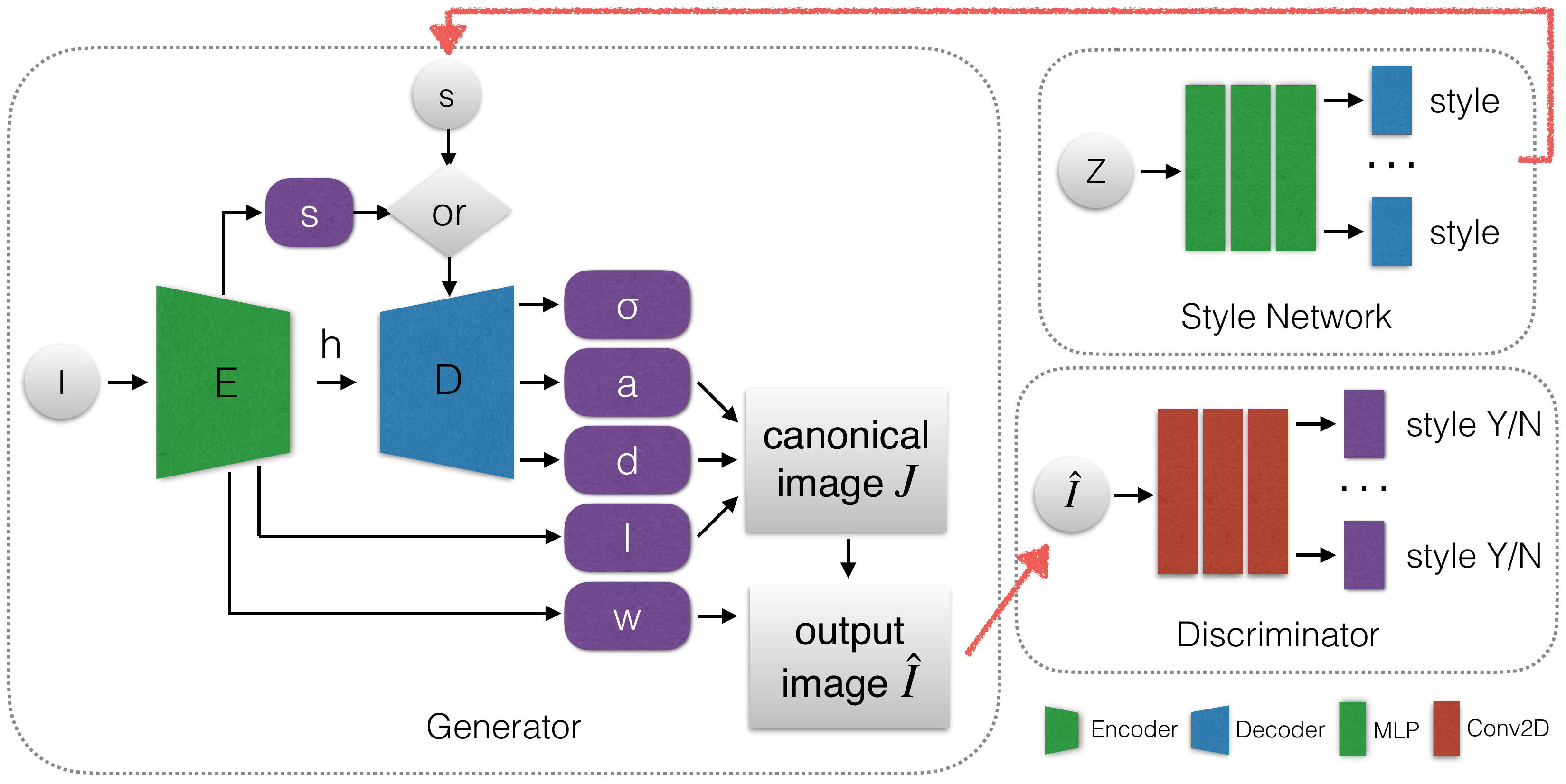}
\caption{Schematic diagram of our unsupervised network architecture. The generator (left) is an autoencoder with heads and rendering blocks that takes as input a 2D image $I$ (and optionally a style code $s$), infers five explicit 3D rendering features (confidence map $\sigma$, albedo $a$, depth $d$, light direction $l$ and viewpoint $w$) and renders the 2D (optionally style transferred) image $\hat{I}$. The style network (right top) maps the latent noise $z$ into style code $s$, promoting style diversity. The discriminator (right bottom) classifies whether the generated image is an instance of a style combination, and is fooled by the generator.
}
\label{fig:main}
\vspace{-0.5cm}
\end{figure*}

\subsection{Neural Network Architecture}
Inspired by StarGAN-v2 \cite{choi2020stargan}, our network architecture consists of three components: a generator $G$ (i.e., an unsupervised explicit 3D reconstruction network), a discriminator $D$, and a style network $S$. These facilitate recovering explicit 3D structure via rendering, promoting the generation of a combination of styles, and generating a specific style from random noise, respectively.
The overall framework is illustrated in Fig.~\ref{fig:main}.
We introduce each component separately.

\textbf{Generator} The generator $G$ takes image $I$ (and optionally style $s$) as input and outputs the image in input (or optionally target) style along with all domain specific styles (e.g., male/female, cat/dog).
It can be seen as the combination of the Generator and the Style encoder in StarGAN-v2 \cite{choi2020stargan}, i.e., the function $G(I,s)$ is generating the style transferred image,
however, in contrast to their work, we also train the Style encoder $E_s(I)$ jointly with the rest of $G$, which queries the domain specific styles.
It is worth noting that all styles mentioned in the paper are domain specific, but this does not mean that each domain has only one style. In each domain, the style is related to an image and different images have different styles.
We denote domain specific style as style for brevity in the rest of the paper.\footnote{For more details on the differences and connections between domain and style, we kindly forward the reader to \cite{choi2020stargan}.}

Our generator $G$ is an unsupervised 3D reconstruction network \cite{wu2020unsupervised}.
We summarize its architecture as follows.
The setup starts with an \emph{autoencoder backbone} and five network \emph{heads}, which together output our five rendering features: the global light direction $l$, the viewpoint $w$, the albedo image $a$, the depth map $d$ and the confidence map $\sigma$); which are then used as input to our \emph{lighting} and \emph{reprojection} blocks to output our style transferred image.
First, given input image $I$, we use our encoder $E$ and two heads to infer the hidden representation $h$, the style code $s=E_s(I)$ and two rendering features ($l$ and $w$).
Optionally, at this point, the style code $s$ can be altered for further control over the styles (e.g., using our style network $S(z)$).
Next, given $h$ and $s$, we apply our decoder $D$ and three heads to compute the other three rendering features ($a$, $d$ and $\sigma$).
Afterwards, given $a$, $d$ and $l$, our lighting block computes the shaded canonical image $J$ as seen from the canonical viewpoint.
Finally, given $d$, $w$ and $J$, our reprojection block changes the viewpoint from canonical to $w$.
We explain the details of each block in the following sections.
For further information see the supplementary material. 
\\
\emph{Autoencoder Backbone and Heads.}
Unlike the structure proposed in \cite{wu2020unsupervised}, which uses separate instances of $E$ and $D$ with untied weights, our method uses weight sharing between these networks to generate all rendering features along with style codes for each domain.
We claim that the shared parameters can help learn useful intrinsic features that contribute to all outputs.
However, we also include untied head branches on top of the shared autoencoder for each output (resulting in an overall partially shared architecture).
To keep enough network capacity, we use convolutional layers in the decoder and in the network heads.
We apply Adaptive Instance Normalization (AdaIN) \cite{huang2017arbitrary, karras2019style} in the decoder and inject style code $s$ into all AdaIN layers to provide scaling and shifting vectors.
\\
\emph{Lighting.}
The shaded canonical image $J$ can be generated from $a$, $d$ and $l$.
First, we compute the normal map $n$ from the depth map $d$ by computing the 3D surface normal for each pixel $(u,v)$, i.e., $n\equiv t^u\times t^v$, where $t^u$ and $t^v$ are the surface tangent vectors along the $u$ and $v$ directions, respectively.
Then, we can get the shaded canonical image $J$ as follows:
\begin{align}
  J_{uv}=a_{uv}\cdot\Bigl[k_s+k_d\max\bigl(0,\langle l, n_{uv} \rangle\bigr)\Bigr],
\end{align}
where $k_s$ and $k_d$ are scalars representing the weights of the ambient and diffuse terms, respectively.\footnote{Further details, e.g., the formulae for $t^u$, $t^v$ and the light direction $l$ can be found in \cite{wu2020unsupervised}.}
\\
\emph{Reprojection.}
We assume a perspective camera with relatively narrow field of view (FOV) of $\theta_{FOV} \approx \SI{10}{\degree}$, since the images are cropped accordingly.
We also assume that the nominal distance between the object and the camera is $\SI{1}{\meter}$.
Then mapping a 3D point $P \in \mathbb{R}^3$ from the reference frame of the camera to a pixel $p=(u,v,1)$ can be expressed as:
\begin{align}
  p\equiv KP,\quad K=\left[   
  \begin{matrix} 
    f\ & 0\ & c_u\\
    0\ & f\ & c_v\\ 
    0\ & 0\ & 1\\ 
  \end{matrix}
\right],     
\end{align}
where $c_u=(W-1)/2$, $c_v=(H-1)/2$ and $f=(W-1)/\bigl[2\tan(\theta_{FOV}/2)\bigr]$.
Since the depth map $d$ associates a depth value $d_{uv}$ to each pixel at position $(u,v)$ in the canonical view, we can get:
\begin{align}
  P=d_{uv}\cdot K^{-1}p.
\end{align}
The viewpoint $w\in \mathbb{R}^6$ is an Euclidean transformation.
$w_{1:3}$ and $w_{4:6}$ represent the rotation angles and the translations along the $x,y,z$ axes, respectively, which we denote by $(R,T)\in SE(3)$.
Then, the warping transformation $A_{d,w}$ of pixels from the canonical view to the actual view is:
\begin{align}
  p'\equiv K(d_{uv}RK^{-1}p+T),
\end{align}
where $p'=(u',v',1)$, and $(u',v')$ is the pixel in the actual view mapped from pixel $(u, v)$ in the canonical view.
Thus, given a canonical image $J$, we output the image $\hat{I}$ as $\hat{I}_{u'v'}=J_{uv}$, where $(u,v)=A_{d,w}^{-1}(u',v')$.


\textbf{Discriminator}
The discriminator $D$ aims to determine whether the input image is a `real' image of its styles or a `generated' image produced by $G$. It consists of $N$ output branches, each of which is a binary classifier in each domain with output dimensions $K=1$ for `real'/`generated' classification.
It is similar to the discriminator network in StarGAN-v2 \cite{choi2020stargan}.

\textbf{Style Network}
The style network $S$ is used to generate a style code $s$ from the latent noise $z$ as $s=S(z)$.
$S$ is a Multi-Layer Perceptron (MLP) with $N$ output branches to provide style codes for all available domains, i.e., $s_y = S_y(z)$, where $S_y$ outputs the style corresponding to the domain $y$.
It is the same as the mapping network proposed in StarGAN-v2 \cite{choi2020stargan}.
By sampling the domain $y$ and the latent noise $z$ randomly, $S$ can generate diverse style codes (as an alternative to obtaining the style code using the encoder as $s=E_s(I)$).


\subsection{Training objectives}

\textbf{Reconstruction and Perceptual Losses}
Besides obtaining the output image $\hat{I}$ as above, we also compute a second output $\hat{I}'$ using the exact same procedure, except for using horizontally flipped depth $d'=flip(d)$ and horizontally flipped albedo $a'=flip(a)$.
We then enforce $I\approx\hat{I}$ and $I\approx\hat{I}'$ using the following combination of losses:
\begin{align}
\begin{split}
  \mathcal{L}_{rec}=&\bigl[\mathcal{L}(I,\hat{I},\sigma)+\lambda_p\cdot\mathcal{L}_p(I,\hat{I},\sigma)\bigr]\\
  &+\lambda_{rec}\cdot\bigl[\mathcal{L}(I,\hat{I}',\sigma')+\lambda_p\cdot\mathcal{L}_p(I,\hat{I}',\sigma')\bigr],
\end{split}
\end{align}
where 
\begin{align}
  \mathcal{L}(I,\hat{I},\sigma)&=-\frac{1}{\lvert\Omega\rvert}\sum_{uv\in\Omega}\ln{\biggl[\frac{1}{\sqrt{2}\sigma_{uv}}}\exp{\Bigl(-\frac{\sqrt{2}\ell_{1,uv}}{\sigma_{uv}}\Bigr)}\biggr],\\
  \mathcal{L}_p(I,\hat{I},\sigma)&=-\frac{1}{\lvert\Omega\rvert}\sum_{uv\in\Omega}\ln{\biggl[\frac{1}{\sqrt{2\pi}\sigma_{uv}}}\exp{\Bigl(-\frac{\ell_{2,p,uv}^2}{2\sigma_{uv}^2}\Bigr)}\biggr]
\end{align}
are negative log-likelihood losses with factorized Laplacian and Gaussian priors, respectively, i.e., $\ell_{1,uv}=\| I_{uv}-\hat{I}_{uv} \|_1$ and $\ell_{2,p,uv}=\| VGG(I_{uv})-VGG(\hat{I}_{uv})\|_2$, where $VGG$ is the \texttt{relu3\_3} feature extractor of VGG16; $\lambda_{rec}$, $\lambda_p$ are balancing factors; $\sigma, \sigma' \in\mathbb{R}^{W\times H}$ are the confidence maps estimating the level of asymmetry for each input pixel.
This implicitly encourages $d\approx d'$, $a\approx a'$, i.e., symmetry in the canonical frame, as proposed in  \cite{wu2020unsupervised}.

\textbf{Adversarial Loss} 
The generator $G$ is trained to fool the discriminator $D$ into thinking that the `generated' image $\hat{I}$ is a `real' instance of target style code $\tilde{s}$ via the following adversarial loss:
\begin{align}
  \mathcal{L}_{adv}=\mathbb{E}_I\biggl[\log D(I)\biggr]+\mathbb{E}_{I,z}\biggl[\log \Bigl(1-D\bigl[G(I,\tilde{s})\bigr]\Bigr)\biggr].
\end{align}
During training, the target style code $\tilde{s}$ is obtained either from the random noise (i.e., $\tilde{s}=S(z)$) or from the input image ($\tilde{s}=E_s(I)$).

\textbf{Style Consistency} 
We add a style consistency loss to enforce the generator $G$ to be able to reconstruct the target style code $\tilde{s}$ from the generated the output image:
\begin{align}
  \mathcal{L}_{sty}=\mathbb{E}_{I,z}\biggl[\Bigl\Vert \tilde{s}-E_s\bigl[G(I,\tilde{s})\bigr]\Bigr\Vert_1\biggr].
\end{align}
Intuitively, this will promote $G$ to utilize $\tilde{s}$.
Different from StarGAN-v2 \cite{choi2020stargan}, instead of using a separate encoder, we use the style encoder $E_s$ of our generator $G$ for obtaining the style code of the output image.

\textbf{Source Image Consistency} 
In line with recent works on adversarial learning  \cite{choi2018stargan, choi2020stargan, kim2017learning, zhu2017unpaired}, we add another cycle consistency loss to enforce $G$ to be able to recover the input image from its style transferred variant:
\begin{align}
  \mathcal{L}_{sou}=\mathbb{E}_{I,z}\biggl[\Bigl\Vert I-G\bigl[G(I,\tilde{s}),E_s(I)\bigr] \Bigr\Vert_1\biggr].
\end{align}
Intuitively, this will promote preserving the original style-invariant characteristics of $I$.
Note that the target style $\tilde{s}$ can be different from the inferred style of the input image $E_s(I)$.

\textbf{Canonical Image Consistency}
In addition, we also enforce $G$ to reconstruct the shaded canonical image $J=G_l(I)$ from itself using its canonical 3D reconstructor $G_l$:
\begin{align}
  \mathcal{L}_{can}=\mathbb{E}_{I}[\Vert G_l(I)-G_l(G_l(I)) \Vert_1].
\end{align}
Intuitively, this helps $G$ to discover the correct canonical frontal view.
During our experiments, we found that this also improves the stability of adversarial training.

\textbf{Style Diversification} 
Finally, we also add a style diversity loss to enforce $G$ to make two style transferred versions of the same input image as different as possible:
\begin{align}
  \mathcal{L}_{sd}=-\mathbb{E}_{I,z_1,z_2}[\Vert G(I,\tilde{s}_1)-G(x,\tilde{s}_2) \Vert_1],
\end{align}
where the target style codes $\tilde{s}_1$ and $\tilde{s}_2$ are either obtained from two random noise samples ($\tilde{s}_i=S(z_i), i\in\{1,2\}$) or from two images from the same domain ($\tilde{s}_i=E_s(I_i), i\in\{1,2\}$).
Intuitively, minimizing this term forces $G$ to explore and learn more diverse images even within the same style domain.

\textbf{Full Objective} 
The total loss function is as follows:
\begin{align}
\begin{split}
\min_{G,S}\max_D \ \mathcal{L}_{rec} + \mathcal{L}_{adv} &+ \lambda_{sty}\cdot\mathcal{L}_{sty} + \lambda_{sou}\cdot\mathcal{L}_{sou} \\
&+ \lambda_{can}\cdot\mathcal{L}_{can} + \lambda_{sd}\cdot\mathcal{L}_{sd},
\end{split}
\end{align}
where $\lambda_{sty}$, $\lambda_{sou}$, $\lambda_{can}$ and $\lambda_{sd}$ are weighting factors.

\begin{table}
\caption{Quantitative experimental results on the BFM dataset. We compare the methods in terms of the Scale-Invariant Depth Error (SIDE) \cite{eigen2014depth} and Mean Angle Deviation (MAD) scores for 3D depth reconstruction, and their respective standard deviations. Winning numbers are highlighted in bold. ${}^\dagger$Numbers here for Wu et al.\ \cite{wu2020unsupervised} are from our reproduction using CUDA 10.2, which is known to perform slightly worse than their published scores using CUDA 9.0.\tablefootnote{See \url{https://github.com/elliottwu/unsup3d/issues/15} for details.}} 

\centering


 %

\label{table:bfm}
\begin{tabular}{lccc}
\toprule
\multicolumn{1}{c}{Method} & \multicolumn{1}{c}{\makecell[c]{Super- \\ vised}} & \multicolumn{1}{c}{SIDE ($\times 10^{-2}$) $\downarrow$} & \multicolumn{1}{c}{MAD (${}^\circ$) $\downarrow$}\\
\midrule
\multirow{2}{*}{\makecell[c]{Wu et al.\ \\ \cite{wu2020unsupervised}${}^\dagger$} } & yes & $0.521{\pm} 0.122$ & $12.36{\pm} 1.13$\\
& no & $0.892{\pm} 0.175$ & $16.98{\pm} 1.07$\\
\midrule
\multirow{2}{*}{Ours} & yes & $\textBF{0.481}{\pm} 0.154$ & $\textBF{10.64}{\pm}  1.29$\\
& no & $\textBF{0.844}{\pm} 0.117$ & $\textBF{14.89}{\pm} 1.33$\\
\bottomrule
\end{tabular}


\vspace{-0.5cm}
\end{table}



\section{Experiments}
\subsection{Datasets}
We evaluated our method on three {}---{} two real and one synthetic {}---{} human face datasets: CelebA-HQ  \cite{karras2017progressive}, 3DFAW  \cite{gross2010multi, jeni2015dense, zhang2013high, zhang2014bp4d} and BFM \cite{paysan20093d}.
CelebA-HQ is a high-quality version of CelebA  \cite{liu2015deep}, which is a large scale human face dataset and includes over 200k images of real human faces in the wild with bounding box annotations.
CelebA-HQ consists of 30k images at $1024 \times 1024$ resolution.
3DFAW contains 23k images with 66 3D face keypoint annotations, which can be used for 3D face alignment and 3D facial landmark localization.
Here, we used 3DFAW to evaluate our 3D prediction performance and compare with baselines at keypoint locations.
Given that in-the-wild datasets lack ground-truth, we used the BFM model to generate a supervised dataset by sampling shapes, poses, textures, and illumination randomly according to the protocol of  \cite{sengupta2018sfsnet}.
We also used images from SUN Database \cite{xiao2010sun} as background and obtain ground truth depth maps for evaluation as proposed in  \cite{wu2020unsupervised}. 

\subsection{Metrics}
We evaluated our method on two problems related to UE3DST, single view 2D to 3D reconstruction and 2D style transfer. We hypothesize that solving both tasks simultaneously regularizes each other.

For 3D reconstruction, we assessed 2D depth map recovery by applying Scale-Invariant Depth Error (SIDE) \cite{eigen2014depth}, Mean Angle Deviation (MAD) and Sum of Depth Correlations (SDC) \cite{moniz2018unsupervised}.
SIDE can be written as:
\begin{align}
\begin{split}
  E_{SIDE}(\Bar{d}, d^*)&=\sqrt{\frac{1}{WH}\sum_{uv}\biggl[\Delta_{uv}^2-\Bigl(\frac{\sum_{uv}\Delta_{uv}}{WH}\Bigr)^2\biggr]}, \\
  \Delta_{uv}&=\log \Bar{d}_{uv}-\log d^*_{uv},
\end{split}
\end{align}
where $d^*$ is the ground-truth depth map and $\Bar{d}$ is the actual view depth map warped from predicted depth map $d$ in the canonical view using the predicted viewpoint.
MAD compares the angles between normals $n^*$ and $n$ computed from $d^*$ and $d$, respectively, as in  \cite{wu2020unsupervised}.
For keypoint depth evaluation, we computed the SDC score between 66 ground truth and predicted frontal facial keypoint locations (i.e., the score is between 0 and 66) as in  \cite{moniz2018unsupervised}.
For image synthesis evaluation, we used Fr\'echet Inception Distance (FID) \cite{heusel2017gans} and Kernel Inception Distance (KID) \cite{binkowski2018demystifying}.

\subsection{Implementation Details}
We split CelebA-HQ into two domains: male and female. For 3DFAW, we roughly cropped images around the head region and used the official training/validation/test splits.
We resized images in all datasets to $64\times64$ in order to fit our method onto the GPUs available to us.
All the experiments were conducted on an 11GB NVIDIA GeForce RTX 2080Ti graphics card with CUDA 10.2 and PyTorch 1.4.0 \cite{paszke2019pytorch}.
We used Adam optimizer \cite{kingma2014adam} with learning rate $\SI{1e-4}{}$, $\beta_1=0.0$, $\beta_2=0.99$ and weight decay $\SI{1e-4}{}$.
We set $\lambda_{rec}=0.5$, $\lambda_p=1$, $\lambda_{sty}=0.5$, $\lambda_{sou}=0.5$, $\lambda_{can}=0.3$, $\lambda_{sd}=0.5$ for all experiments.
We used batch size 16 and train 100k iterations in total.
To stabilize the adversarial training process, first, we trained the generator $G$ alone for the first 20k iterations, and then trained the whole network jointly for the remaining 80k iterations.

\begin{table}
\begin{center}
\caption{Quantitative experimental results on the 3DFAW dataset. We compare the methods in terms of the Sum of Depth Correlations (SDC) \cite{moniz2018unsupervised} score for keypoint depth. Winning numbers are highlighted in bold for each category. ${}^\dagger$Numbers here for Wu et al.\ \cite{wu2020unsupervised} are from our reproduction using CUDA 10.2, which is known to perform slightly worse than their published scores using CUDA 9.0.\tablefootnote{See \url{https://github.com/elliottwu/unsup3d/issues/15} for details.} Besides, the numbers may further differ due to using our custom implementation for evaluation.\tablefootnote{See \url{https://github.com/elliottwu/unsup3d/issues/10} for details.}}
\label{table:3DFAW}
\begin{tabular}{lcc}
\toprule
\multicolumn{1}{c}{Method} & \multicolumn{1}{c}{Supervised} & \multicolumn{1}{c}{SDC $\uparrow$}\\
\midrule
Ground truth & N/A & $66.00$\\
\midrule
AIGN \cite{tung2017adversarial} & yes \cite{moniz2018unsupervised} & $50.81$\\
DepthNetGAN\cite{moniz2018unsupervised} & yes \cite{moniz2018unsupervised} & $\textBF{58.68}$ \\
MOFA \cite{tewari2017mofa} & yes, model-based \cite{moniz2018unsupervised} & $15.97$\\
\midrule
DepthNet \cite{moniz2018unsupervised} & no, trained on \cite{moniz2018unsupervised} & $26.32$ \\
DepthNet \cite{moniz2018unsupervised} & no & $35.77$ \\
Wu et al.\ \cite{wu2020unsupervised}${}^\dagger$ & no & $41.52$\\
Ours & no & $\textBF{44.86}$\\
\bottomrule
\end{tabular}
\end{center}
\vspace{-0.5cm}
\end{table}

\begin{table}
\begin{center}
\caption{Quantitative results on the CelebA-HQ $64\times 64$ dataset. We compare the methods in terms of the Fr\'echet Inception Distance (FID) \cite{heusel2017gans} and Kernel Inception Distance (KID) \cite{binkowski2018demystifying} scores for image synthesis. Winning numbers are highlighted in bold.}
\label{table:fid}
\begin{tabular}{lcccc}
\toprule
& & \multicolumn{1}{c}{Img.} &  & \multicolumn{1}{c}{KID} \\
\multicolumn{1}{c}{Method} & \multicolumn{1}{c}{3D} & \multicolumn{1}{c}{Cond.} & \multicolumn{1}{c}{FID $\downarrow$} & \multicolumn{1}{c}{($\times 100$) $\downarrow$}\\
\midrule
HoloGAN  \cite{nguyen2019hologan} & aware & no & \phantom{0}N/A & $2.87$\\
$\pi$-GAN  \cite{chan2021pi} & NeRF & no & $\phantom{0}\textBF{5.15}$ & $\textBF{0.09}$\\
Pix2NeRF \cite{cai2022pix2nerf} & NeRF & no & $\phantom{0}6.25$ & $0.16$\\
\midrule
Pix2NeRF \cite{cai2022pix2nerf} & NeRF & yes & $24.64$ & $1.93$\\
Ours & explic. & yes & $\phantom{0}\textBF{8.19}$ & $\textBF{0.78}$\\
\bottomrule
\end{tabular}
\end{center}
\vspace{-0.4cm}
\end{table}

\begin{figure*}
\centering
\includegraphics[width=0.85\textwidth]{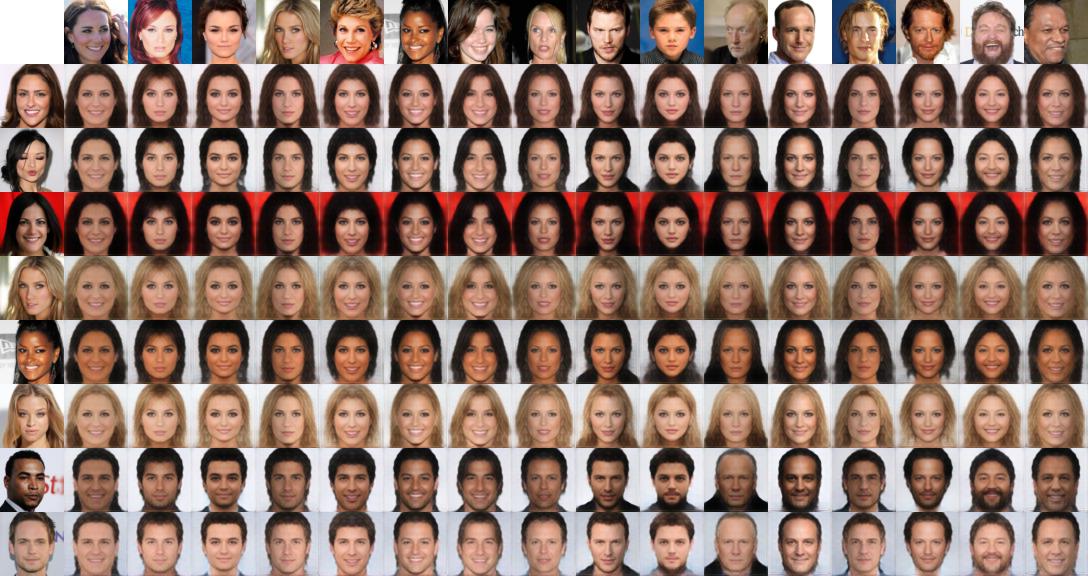}
\caption{Reference-guided image synthesis results on CelebA-HQ. First row is source images and first column is reference images. We can see that our method can generate realistic images with the style of the reference image based on the source image.}
\label{fig:ref}
\end{figure*}

\begin{figure*}
\centering
\includegraphics[width=0.85\textwidth]{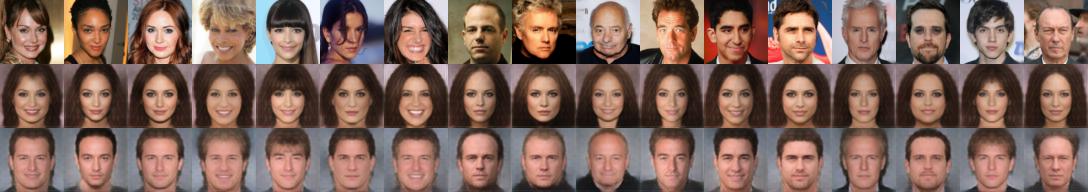}
\caption{Latent noise-guided image synthesis results on CelebA-HQ. First row is source images and we generate $z$ randomly as input. We can see that our method can generate realistic images by controlling the style through the latent space.}
\label{fig:lat}
\end{figure*}

\begin{figure*}
\centering
\includegraphics[width=0.60\textwidth]{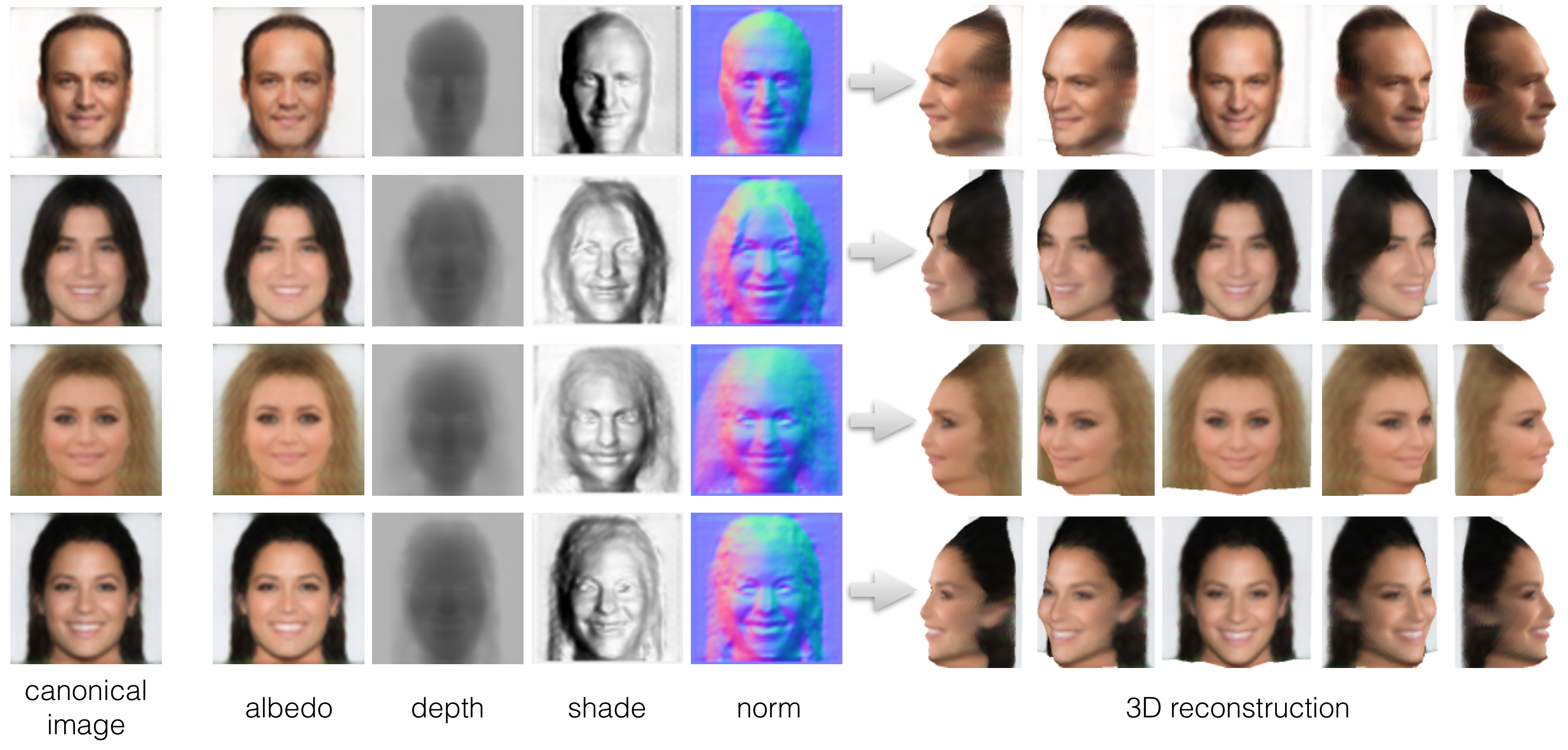}
\caption{3D reconstruction from single view image visualization. Our method can effectively recover the 3D structure of style-transferred images.}
\label{fig:vis}
\end{figure*}

\begin{figure*}
\centering
\includegraphics[width=0.85\textwidth]{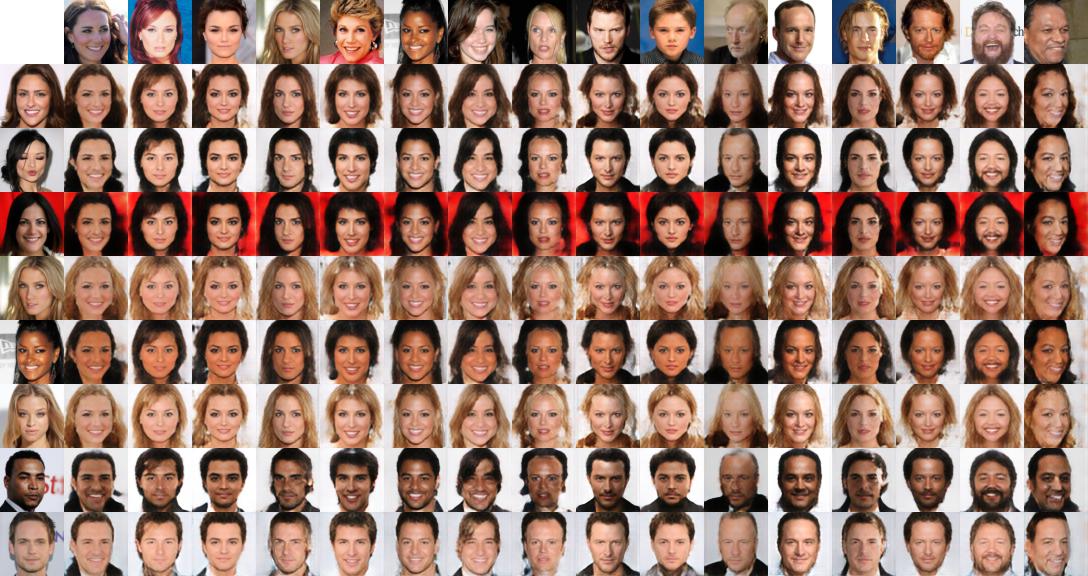}
\caption{Ablation study on canonical image consistency. We found that canonical image consistency plays a crucial role in the quality of the generated images.}
\label{fig:ablation2}
\end{figure*}

\begin{table}
\begin{center}
\caption{Quantitative ablation study results on the BFM dataset. We compare the methods in terms of the Scale-Invariant Depth Error (SIDE) \cite{eigen2014depth} and Mean Angle Deviation (MAD) scores for 3D depth reconstruction, and their respective standard deviations. Winning numbers are highlighted in bold.}
\label{table:ablation1}
\begin{tabular}{lcc}
\toprule
\multicolumn{1}{c}{Method} & \multicolumn{1}{c}{SIDE ($\times 10^{-2}$) $\downarrow$} & \multicolumn{1}{c}{MAD (${}^\circ$) $\downarrow$}\\
\midrule
Ours, w/o albedo flip & $3.148{\pm} 0.322$ & $41.21{\pm} 2.01$\\
Ours, w/o depth flip & $1.463{\pm} 0.241$ & $29.87{\pm} 2.10$\\
Ours, w/o conf. map & $0.885{\pm} 0.131$ & $15.16{\pm} 1.23$\\
Ours, full model & $\textBF{0.844}{\pm} 0.117$ & $\textBF{14.89}{\pm} 1.33$\\
\bottomrule
\end{tabular}
\end{center}
\vspace{-0.3cm}
\end{table}

\subsection{Results}
We used the BFM dataset to compare the 2D depth map reconstruction quality obtained by our method with the baseline of Wu et al.\ \cite{wu2020unsupervised}, as well as their respective supervised rendering-free equivalents that regress the ground-truth depth maps directly using an $\ell_1$ loss.
We retrained the baseline model from scratch using CUDA 10.2, since their published results were obtained with CUDA 9.0.
The results are shown in Table~\ref{table:bfm}.
We can see that our method outperforms the baseline, which confirms that our shared encoder-decoder and adversarial losses together yield a better instance specific 3D representation.

Next, we compared the 3D keypoint depth reconstruction quality with baselines Wu et al.\ \cite{wu2020unsupervised}, AIGN \cite{tung2017adversarial}, MOFA \cite{tewari2017mofa} and DepthNet \cite{moniz2018unsupervised}.
Note that AIGN, MOFA and DepthNet each have unfair advantage over our method, as the former two are supervised, whereas the latter takes 2D keypoint locations as input.
For all algorithms except DepthNet, we the computed the full 2D depth map $d$ and sampled it at keypoint locations, in order to compute the SDC score against the ground truth.
We reimplemented the evaluation code based upon \cite{wu2020unsupervised}, since they did not release this part of their code, causing further potential mismatch between our replicated and their published performance, besides CUDA version differences. 
The results are collected in Table~\ref{table:3DFAW}.
We found that our model performs much better than other model-based or unsupervised methods, further confirming the advantages of using a tied weights autoencoder and style transfer regularization.

Lastly, for image synthesis quality evaluation, we computed FID and KID on the CelebA-HQ dataset, and compared with 3D aware and NeRF-integrated GAN baselines: HoloGAN, $\pi$-GAN and two variants of Pix2NeRF.
Again, note that some of these baselines are image unconditional, having an unfair advantage over our conditional procedure, since feeding image inputs limits the variability and diversity of output images, affecting both the FID and KID scores.
The results are summarized in Table~\ref{table:fid}.
We can see that our method  outperformed the image conditional Pix2NeRF by a large margin; and is approaching the level of unconditional methods, while having explicit 3D reconstruction ability.
We also provide qualitative examples to visualize our image synthesis results using style codes $s$ both from reference images (Fig.~\ref{fig:ref}) and from noise vectors $z$ (Fig.~\ref{fig:lat}).
Fig.~\ref{fig:vis} depicts the recovered 3D structure of some style transferred images, which are spectacular for inferring from a single viewpoint only.

\subsection{Ablation Study}
We performed ablation experiments on the BFM dataset to confirm that the model exploits symmetry/asymmetry information and canonical consistency properly.

First, we quantitatively evaluated the efficiency of the flip operation and the confidence map, as in \cite{wu2020unsupervised}.
As shown in Table~\ref{table:ablation1}, we can see that performance degrades drastically without applying flipping on depth or albedo map.
The lack of a confidence map affected the scores less, but was still worse than our full model.

Second, we show qualitative results for our method without canonical image consistency loss in Fig.~\ref{fig:ablation2}.
We can see that the model is not able learn good frontal view canonical images without the canonical image consistency loss, which confirms the rationality and necessity of this term.


\section{Conclusions}
In this paper, we proposed an end-to-end unsupervised network for 2D to explicit 3D style transfer.
The method combines existing approaches for two related problems: the ill-posedness of 2D to 3D reconstruction is alleviated by utilizing albedo and depth symmetry, whereas adversarial training is stabilized by our cycle consistency losses.
Our quantitative and qualitative experiments showed that our scheme achieves performance better or comparable to prior works on the two tasks, while solving both at the same time.
Future work may consider using more powerful GAN architectures, richer shading models (e.g., shadows, specularity), more complex objects and priors beyond faces and symmetry (e.g., multiple canonical views) and integration with other explicit 3D representations (e.g., meshes).
{\small
\bibliographystyle{ieee_fullname}
\bibliography{egbib}

\begin{thebibliography}{10}\itemsep=-1pt

\bibitem{agrawal2015learning}
Pulkit Agrawal, Joao Carreira, and Jitendra Malik.
\newblock Learning to see by moving.
\newblock In {\em Proceedings of the IEEE international conference on computer
  vision}, pages 37--45, 2015.

\bibitem{aizawa2021agnostic}
Hiroaki Aizawa, Hirokatsu Kataoka, Yutaka Satoh, and Kunihito Kato.
\newblock agnostic image rendering.
\newblock In {\em Proceedings of the IEEE/CVF Winter Conference on Applications
  of Computer Vision}, pages 3803--3812, 2021.

\bibitem{berthelot2017began}
David Berthelot, Thomas Schumm, and Luke Metz.
\newblock Began: Boundary equilibrium generative adversarial networks.
\newblock {\em arXiv preprint arXiv:1703.10717}, 2017.

\bibitem{binkowski2018demystifying}
Miko{\l}aj Bi{\'n}kowski, Danica~J Sutherland, Michael Arbel, and Arthur
  Gretton.
\newblock Demystifying mmd gans.
\newblock {\em arXiv preprint arXiv:1801.01401}, 2018.

\bibitem{bregler2000recovering}
Christoph Bregler, Aaron Hertzmann, and Henning Biermann.
\newblock Recovering non-rigid 3d shape from image streams.
\newblock In {\em Proceedings IEEE Conference on Computer Vision and Pattern
  Recognition. CVPR 2000 (Cat. No. PR00662)}, volume~2, pages 690--696. IEEE,
  2000.

\bibitem{brock2018large}
Andrew Brock, Jeff Donahue, and Karen Simonyan.
\newblock Large scale gan training for high fidelity natural image synthesis.
\newblock {\em arXiv preprint arXiv:1809.11096}, 2018.

\bibitem{cai2022pix2nerf}
Shengqu Cai, Anton Obukhov, Dengxin Dai, and Luc Van~Gool.
\newblock Pix2nerf: Unsupervised conditional pi-gan for single image to neural
  radiance fields translation.
\newblock {\em arXiv preprint arXiv:2202.13162}, 2022.

\bibitem{chan2021pi}
Eric~R Chan, Marco Monteiro, Petr Kellnhofer, Jiajun Wu, and Gordon Wetzstein.
\newblock pi-gan: Periodic implicit generative adversarial networks for
  3d-aware image synthesis.
\newblock In {\em Proceedings of the IEEE/CVF conference on computer vision and
  pattern recognition}, pages 5799--5809, 2021.

\bibitem{chen2022sofgan}
Anpei Chen, Ruiyang Liu, Ling Xie, Zhang Chen, Hao Su, and Jingyi Yu.
\newblock Sofgan: A portrait image generator with dynamic styling.
\newblock {\em ACM Transactions on Graphics (TOG)}, 41(1):1--26, 2022.

\bibitem{chen2019unsupervised}
Ching-Hang Chen, Ambrish Tyagi, Amit Agrawal, Dylan Drover, Stefan Stojanov,
  and James~M Rehg.
\newblock Unsupervised 3d pose estimation with geometric self-supervision.
\newblock In {\em Proceedings of the IEEE/CVF Conference on Computer Vision and
  Pattern Recognition}, pages 5714--5724, 2019.

\bibitem{choi2018stargan}
Yunjey Choi, Minje Choi, Munyoung Kim, Jung-Woo Ha, Sunghun Kim, and Jaegul
  Choo.
\newblock Stargan: Unified generative adversarial networks for multi-domain
  image-to-image translation.
\newblock In {\em Proceedings of the IEEE conference on computer vision and
  pattern recognition}, pages 8789--8797, 2018.

\bibitem{choi2020stargan}
Yunjey Choi, Youngjung Uh, Jaejun Yoo, and Jung-Woo Ha.
\newblock Stargan v2: Diverse image synthesis for multiple domains.
\newblock In {\em Proceedings of the IEEE/CVF conference on computer vision and
  pattern recognition}, pages 8188--8197, 2020.

\bibitem{deng2020disentangled}
Yu Deng, Jiaolong Yang, Dong Chen, Fang Wen, and Xin Tong.
\newblock Disentangled and controllable face image generation via 3d
  imitative-contrastive learning.
\newblock In {\em Proceedings of the IEEE/CVF conference on computer vision and
  pattern recognition}, pages 5154--5163, 2020.

\bibitem{deng2021gram}
Yu Deng, Jiaolong Yang, Jianfeng Xiang, and Xin Tong.
\newblock Gram: Generative radiance manifolds for 3d-aware image generation.
\newblock {\em arXiv preprint arXiv:2112.08867}, 2021.

\bibitem{eigen2014depth}
David Eigen, Christian Puhrsch, and Rob Fergus.
\newblock Depth map prediction from a single image using a multi-scale deep
  network.
\newblock {\em Advances in neural information processing systems}, 27, 2014.

\bibitem{geng2018warp}
Jiahao Geng, Tianjia Shao, Youyi Zheng, Yanlin Weng, and Kun Zhou.
\newblock Warp-guided gans for single-photo facial animation.
\newblock {\em ACM Transactions on Graphics (TOG)}, 37(6):1--12, 2018.

\bibitem{godard2017unsupervised}
Cl{\'e}ment Godard, Oisin Mac~Aodha, and Gabriel~J Brostow.
\newblock Unsupervised monocular depth estimation with left-right consistency.
\newblock In {\em Proceedings of the IEEE conference on computer vision and
  pattern recognition}, pages 270--279, 2017.

\bibitem{goodfellow2014generative}
Ian Goodfellow, Jean Pouget-Abadie, Mehdi Mirza, Bing Xu, David Warde-Farley,
  Sherjil Ozair, Aaron Courville, and Yoshua Bengio.
\newblock Generative adversarial nets.
\newblock {\em Advances in neural information processing systems}, 27, 2014.

\bibitem{gross2010multi}
Ralph Gross, Iain Matthews, Jeffrey Cohn, Takeo Kanade, and Simon Baker.
\newblock Multi-pie.
\newblock {\em Image and vision computing}, 28(5):807--813, 2010.

\bibitem{gu2021stylenerf}
Jiatao Gu, Lingjie Liu, Peng Wang, and Christian Theobalt.
\newblock Stylenerf: A style-based 3d-aware generator for high-resolution image
  synthesis.
\newblock {\em arXiv preprint arXiv:2110.08985}, 2021.

\bibitem{heusel2017gans}
Martin Heusel, Hubert Ramsauer, Thomas Unterthiner, Bernhard Nessler, and Sepp
  Hochreiter.
\newblock Gans trained by a two time-scale update rule converge to a local nash
  equilibrium.
\newblock {\em Advances in neural information processing systems}, 30, 2017.

\bibitem{huang2017arbitrary}
Xun Huang and Serge Belongie.
\newblock Arbitrary style transfer in real-time with adaptive instance
  normalization.
\newblock In {\em Proceedings of the IEEE international conference on computer
  vision}, pages 1501--1510, 2017.

\bibitem{pix2pix2017}
Phillip Isola, Jun-Yan Zhu, Tinghui Zhou, and Alexei~A Efros.
\newblock Image-to-image translation with conditional adversarial networks.
\newblock {\em CVPR}, 2017.

\bibitem{jeni2015dense}
L{\'a}szl{\'o}~A Jeni, Jeffrey~F Cohn, and Takeo Kanade.
\newblock Dense 3d face alignment from 2d videos in real-time.
\newblock In {\em 2015 11th IEEE international conference and workshops on
  automatic face and gesture recognition (FG)}, volume~1, pages 1--8. IEEE,
  2015.

\bibitem{jung2009novel}
Yong~Ju Jung, Aron Baik, Jiwon Kim, and Dusik Park.
\newblock A novel 2d-to-3d conversion technique based on relative height-depth
  cue.
\newblock In {\em Stereoscopic Displays and Applications XX}, volume 7237,
  pages 589--596. SPIE, 2009.

\bibitem{kanazawa2018learning}
Angjoo Kanazawa, Shubham Tulsiani, Alexei~A Efros, and Jitendra Malik.
\newblock Learning category-specific mesh reconstruction from image
  collections.
\newblock In {\em Proceedings of the European Conference on Computer Vision
  (ECCV)}, pages 371--386, 2018.

\bibitem{karras2017progressive}
Tero Karras, Timo Aila, Samuli Laine, and Jaakko Lehtinen.
\newblock Progressive growing of gans for improved quality, stability, and
  variation.
\newblock {\em arXiv preprint arXiv:1710.10196}, 2017.

\bibitem{karras2019style}
Tero Karras, Samuli Laine, and Timo Aila.
\newblock A style-based generator architecture for generative adversarial
  networks.
\newblock In {\em Proceedings of the IEEE/CVF conference on computer vision and
  pattern recognition}, pages 4401--4410, 2019.

\bibitem{kim2017learning}
Taeksoo Kim, Moonsu Cha, Hyunsoo Kim, Jung~Kwon Lee, and Jiwon Kim.
\newblock Learning to discover cross-domain relations with generative
  adversarial networks.
\newblock In {\em International conference on machine learning}, pages
  1857--1865. PMLR, 2017.

\bibitem{kingma2014adam}
Diederik~P Kingma and Jimmy Ba.
\newblock Adam: A method for stochastic optimization.
\newblock {\em arXiv preprint arXiv:1412.6980}, 2014.

\bibitem{kossaifi2018gagan}
Jean Kossaifi, Linh Tran, Yannis Panagakis, and Maja Pantic.
\newblock Gagan: Geometry-aware generative adversarial networks.
\newblock In {\em Proceedings of the IEEE conference on computer vision and
  pattern recognition}, pages 878--887, 2018.

\bibitem{liu2015deep}
Ziwei Liu, Ping Luo, Xiaogang Wang, and Xiaoou Tang.
\newblock Deep learning face attributes in the wild.
\newblock In {\em Proceedings of the IEEE international conference on computer
  vision}, pages 3730--3738, 2015.

\bibitem{lorensen1987marching}
William~E Lorensen and Harvey~E Cline.
\newblock Marching cubes: A high resolution 3d surface construction algorithm.
\newblock {\em ACM siggraph computer graphics}, 21(4):163--169, 1987.

\bibitem{mildenhall2020nerf}
Ben Mildenhall, Pratul~P Srinivasan, Matthew Tancik, Jonathan~T Barron, Ravi
  Ramamoorthi, and Ren Ng.
\newblock Nerf: Representing scenes as neural radiance fields for view
  synthesis.
\newblock In {\em European conference on computer vision}, pages 405--421.
  Springer, 2020.

\bibitem{moniz2018unsupervised}
Joel Ruben~Antony Moniz, Christopher Beckham, Simon Rajotte, Sina Honari, and
  Chris Pal.
\newblock Unsupervised depth estimation, 3d face rotation and replacement.
\newblock {\em Advances in neural information processing systems}, 31, 2018.

\bibitem{mukherjee1995shape}
Dipti~Prasad Mukherjee, Andrew~Peter Zisserman, Michael Brady, and FT Smith.
\newblock Shape from symmetry: Detecting and exploiting symmetry in affine
  images.
\newblock {\em Philosophical Transactions of the Royal Society of London.
  Series A: Physical and Engineering Sciences}, 351(1695):77--106, 1995.

\bibitem{navaneet2020image}
KL Navaneet, Ansu Mathew, Shashank Kashyap, Wei-Chih Hung, Varun Jampani, and
  R~Venkatesh Babu.
\newblock From image collections to point clouds with self-supervised shape and
  pose networks.
\newblock In {\em Proceedings of the IEEE/CVF Conference on Computer Vision and
  Pattern Recognition}, pages 1132--1140, 2020.

\bibitem{nguyen2019hologan}
Thu Nguyen-Phuoc, Chuan Li, Lucas Theis, Christian Richardt, and Yong-Liang
  Yang.
\newblock Hologan: Unsupervised learning of 3d representations from natural
  images.
\newblock In {\em Proceedings of the IEEE/CVF International Conference on
  Computer Vision}, pages 7588--7597, 2019.

\bibitem{nguyen2018rendernet}
Thu~H Nguyen-Phuoc, Chuan Li, Stephen Balaban, and Yongliang Yang.
\newblock Rendernet: A deep convolutional network for differentiable rendering
  from 3d shapes.
\newblock {\em Advances in Neural Information Processing Systems}, 31, 2018.

\bibitem{niemeyer2021giraffe}
Michael Niemeyer and Andreas Geiger.
\newblock Giraffe: Representing scenes as compositional generative neural
  feature fields.
\newblock In {\em Proceedings of the IEEE/CVF Conference on Computer Vision and
  Pattern Recognition}, pages 11453--11464, 2021.

\bibitem{novotny2017learning}
David Novotny, Diane Larlus, and Andrea Vedaldi.
\newblock Learning 3d object categories by looking around them.
\newblock In {\em Proceedings of the IEEE International Conference on Computer
  Vision}, pages 5218--5227, 2017.

\bibitem{novotny2022keytr}
David Novotny, Ignacio Rocco, Samarth Sinha, Alexandre Carlier, Gael
  Kerchenbaum, Roman Shapovalov, Nikita Smetanin, Natalia Neverova, Benjamin
  Graham, and Andrea Vedaldi.
\newblock Keytr: Keypoint transporter for 3d reconstruction of deformable
  objects in videos.
\newblock In {\em Proceedings of the IEEE/CVF Conference on Computer Vision and
  Pattern Recognition}, pages 5595--5604, 2022.

\bibitem{paszke2019pytorch}
Adam Paszke, Sam Gross, Francisco Massa, Adam Lerer, James Bradbury, Gregory
  Chanan, Trevor Killeen, Zeming Lin, Natalia Gimelshein, Luca Antiga, et~al.
\newblock Pytorch: An imperative style, high-performance deep learning library.
\newblock {\em Advances in neural information processing systems}, 32, 2019.

\bibitem{paysan20093d}
Pascal Paysan, Reinhard Knothe, Brian Amberg, Sami Romdhani, and Thomas Vetter.
\newblock A 3d face model for pose and illumination invariant face recognition.
\newblock In {\em 2009 sixth IEEE international conference on advanced video
  and signal based surveillance}, pages 296--301. Ieee, 2009.

\bibitem{ranjan2018generating}
Anurag Ranjan, Timo Bolkart, Soubhik Sanyal, and Michael~J Black.
\newblock Generating 3d faces using convolutional mesh autoencoders.
\newblock In {\em Proceedings of the European Conference on Computer Vision
  (ECCV)}, pages 704--720, 2018.

\bibitem{rhodin2018unsupervised}
Helge Rhodin, Mathieu Salzmann, and Pascal Fua.
\newblock Unsupervised geometry-aware representation for 3d human pose
  estimation.
\newblock In {\em Proceedings of the European conference on computer vision
  (ECCV)}, pages 750--767, 2018.

\bibitem{sahasrabudhe2019lifting}
Mihir Sahasrabudhe, Zhixin Shu, Edward Bartrum, Riza Alp~Guler, Dimitris
  Samaras, and Iasonas Kokkinos.
\newblock Lifting autoencoders: Unsupervised learning of a fully-disentangled
  3d morphable model using deep non-rigid structure from motion.
\newblock In {\em Proceedings of the IEEE/CVF International Conference on
  Computer Vision Workshops}, pages 0--0, 2019.

\bibitem{schwarz2020graf}
Katja Schwarz, Yiyi Liao, Michael Niemeyer, and Andreas Geiger.
\newblock Graf: Generative radiance fields for 3d-aware image synthesis.
\newblock {\em Advances in Neural Information Processing Systems},
  33:20154--20166, 2020.

\bibitem{sengupta2018sfsnet}
Soumyadip Sengupta, Angjoo Kanazawa, Carlos~D Castillo, and David~W Jacobs.
\newblock Sfsnet: Learning shape, reflectance and illuminance of facesin the
  wild'.
\newblock In {\em Proceedings of the IEEE conference on computer vision and
  pattern recognition}, pages 6296--6305, 2018.

\bibitem{shi2021lifting}
Yichun Shi, Divyansh Aggarwal, and Anil~K Jain.
\newblock Lifting 2d stylegan for 3d-aware face generation.
\newblock In {\em Proceedings of the IEEE/CVF Conference on Computer Vision and
  Pattern Recognition}, pages 6258--6266, 2021.

\bibitem{shin2018pixels}
Daeyun Shin, Charless~C Fowlkes, and Derek Hoiem.
\newblock Pixels, voxels, and views: A study of shape representations for
  single view 3d object shape prediction.
\newblock In {\em Proceedings of the IEEE conference on computer vision and
  pattern recognition}, pages 3061--3069, 2018.

\bibitem{sidhu2020neural}
Vikramjit Sidhu, Edgar Tretschk, Vladislav Golyanik, Antonio Agudo, and
  Christian Theobalt.
\newblock Neural dense non-rigid structure from motion with latent space
  constraints.
\newblock In {\em European Conference on Computer Vision}, pages 204--222.
  Springer, 2020.

\bibitem{sitzmann2019deepvoxels}
Vincent Sitzmann, Justus Thies, Felix Heide, Matthias Nie{\ss}ner, Gordon
  Wetzstein, and Michael Zollhofer.
\newblock Deepvoxels: Learning persistent 3d feature embeddings.
\newblock In {\em Proceedings of the IEEE/CVF Conference on Computer Vision and
  Pattern Recognition}, pages 2437--2446, 2019.

\bibitem{sun2021neuralrecon}
Jiaming Sun, Yiming Xie, Linghao Chen, Xiaowei Zhou, and Hujun Bao.
\newblock Neuralrecon: Real-time coherent 3d reconstruction from monocular
  video.
\newblock In {\em Proceedings of the IEEE/CVF Conference on Computer Vision and
  Pattern Recognition}, pages 15598--15607, 2021.

\bibitem{suwajanakorn2018discovery}
Supasorn Suwajanakorn, Noah Snavely, Jonathan~J Tompson, and Mohammad Norouzi.
\newblock Discovery of latent 3d keypoints via end-to-end geometric reasoning.
\newblock {\em Advances in neural information processing systems}, 31, 2018.

\bibitem{szabo2019unsupervised}
Attila Szab{\'o}, Givi Meishvili, and Paolo Favaro.
\newblock Unsupervised generative 3d shape learning from natural images.
\newblock {\em arXiv preprint arXiv:1910.00287}, 2019.

\bibitem{tewari2017mofa}
Ayush Tewari, Michael Zollhofer, Hyeongwoo Kim, Pablo Garrido, Florian Bernard,
  Patrick Perez, and Christian Theobalt.
\newblock Mofa: Model-based deep convolutional face autoencoder for
  unsupervised monocular reconstruction.
\newblock In {\em Proceedings of the IEEE International Conference on Computer
  Vision Workshops}, pages 1274--1283, 2017.

\bibitem{thewlis2018modelling}
James Thewlis, Hakan Bilen, and Andrea Vedaldi.
\newblock Modelling and unsupervised learning of symmetric deformable object
  categories.
\newblock {\em Advances in Neural Information Processing Systems}, 31, 2018.

\bibitem{thrun2005shape}
Sebastian Thrun and Ben Wegbreit.
\newblock Shape from symmetry.
\newblock In {\em Tenth IEEE International Conference on Computer Vision
  (ICCV'05) Volume 1}, volume~2, pages 1824--1831. IEEE, 2005.

\bibitem{tung2017adversarial}
Hsiao-Yu~Fish Tung, Adam~W Harley, William Seto, and Katerina Fragkiadaki.
\newblock Adversarial inverse graphics networks: Learning 2d-to-3d lifting and
  image-to-image translation from unpaired supervision.
\newblock In {\em 2017 IEEE International Conference on Computer Vision
  (ICCV)}, pages 4364--4372. IEEE, 2017.

\bibitem{ullman1979interpretation}
Shimon Ullman.
\newblock The interpretation of structure from motion.
\newblock {\em Proceedings of the Royal Society of London. Series B. Biological
  Sciences}, 203(1153):405--426, 1979.

\bibitem{wang2018learning}
Chaoyang Wang, Jos{\'e}~Miguel Buenaposada, Rui Zhu, and Simon Lucey.
\newblock Learning depth from monocular videos using direct methods.
\newblock In {\em Proceedings of the IEEE conference on computer vision and
  pattern recognition}, pages 2022--2030, 2018.

\bibitem{wang2018pixel2mesh}
Nanyang Wang, Yinda Zhang, Zhuwen Li, Yanwei Fu, Wei Liu, and Yu-Gang Jiang.
\newblock Pixel2mesh: Generating 3d mesh models from single rgb images.
\newblock In {\em Proceedings of the European conference on computer vision
  (ECCV)}, pages 52--67, 2018.

\bibitem{wang2019pseudo}
Yan Wang, Wei-Lun Chao, Divyansh Garg, Bharath Hariharan, Mark Campbell, and
  Kilian~Q Weinberger.
\newblock Pseudo-lidar from visual depth estimation: Bridging the gap in 3d
  object detection for autonomous driving.
\newblock In {\em Proceedings of the IEEE/CVF Conference on Computer Vision and
  Pattern Recognition}, pages 8445--8453, 2019.

\bibitem{wu2016learning}
Jiajun Wu, Chengkai Zhang, Tianfan Xue, Bill Freeman, and Josh Tenenbaum.
\newblock Learning a probabilistic latent space of object shapes via 3d
  generative-adversarial modeling.
\newblock {\em Advances in neural information processing systems}, 29, 2016.

\bibitem{wu2020unsupervised}
Shangzhe Wu, Christian Rupprecht, and Andrea Vedaldi.
\newblock Unsupervised learning of probably symmetric deformable 3d objects
  from images in the wild.
\newblock In {\em Proceedings of the IEEE/CVF Conference on Computer Vision and
  Pattern Recognition}, pages 1--10, 2020.

\bibitem{xiao2010sun}
Jianxiong Xiao, James Hays, Krista~A Ehinger, Aude Oliva, and Antonio Torralba.
\newblock Sun database: Large-scale scene recognition from abbey to zoo.
\newblock In {\em 2010 IEEE computer society conference on computer vision and
  pattern recognition}, pages 3485--3492. IEEE, 2010.

\bibitem{zhang2019self}
Han Zhang, Ian Goodfellow, Dimitris Metaxas, and Augustus Odena.
\newblock Self-attention generative adversarial networks.
\newblock In {\em International conference on machine learning}, pages
  7354--7363. PMLR, 2019.

\bibitem{zhang1999shape}
Ruo Zhang, Ping-Sing Tsai, James~Edwin Cryer, and Mubarak Shah.
\newblock Shape-from-shading: a survey.
\newblock {\em IEEE transactions on pattern analysis and machine intelligence},
  21(8):690--706, 1999.

\bibitem{zhang2013high}
Xing Zhang, Lijun Yin, Jeffrey~F Cohn, Shaun Canavan, Michael Reale, Andy
  Horowitz, and Peng Liu.
\newblock A high-resolution spontaneous 3d dynamic facial expression database.
\newblock In {\em 2013 10th IEEE international conference and workshops on
  automatic face and gesture recognition (FG)}, pages 1--6. IEEE, 2013.

\bibitem{zhang2014bp4d}
Xing Zhang, Lijun Yin, Jeffrey~F Cohn, Shaun Canavan, Michael Reale, Andy
  Horowitz, Peng Liu, and Jeffrey~M Girard.
\newblock Bp4d-spontaneous: a high-resolution spontaneous 3d dynamic facial
  expression database.
\newblock {\em Image and Vision Computing}, 32(10):692--706, 2014.

\bibitem{zhang2020image}
Yuxuan Zhang, Wenzheng Chen, Huan Ling, Jun Gao, Yinan Zhang, Antonio Torralba,
  and Sanja Fidler.
\newblock Image gans meet differentiable rendering for inverse graphics and
  interpretable 3d neural rendering.
\newblock {\em arXiv preprint arXiv:2010.09125}, 2020.

\bibitem{zhou2017unsupervised}
Tinghui Zhou, Matthew Brown, Noah Snavely, and David~G Lowe.
\newblock Unsupervised learning of depth and ego-motion from video.
\newblock In {\em Proceedings of the IEEE conference on computer vision and
  pattern recognition}, pages 1851--1858, 2017.

\bibitem{CycleGAN2017}
Jun-Yan Zhu, Taesung Park, Phillip Isola, and Alexei~A Efros.
\newblock Unpaired image-to-image translation using cycle-consistent
  adversarial networks.
\newblock In {\em Computer Vision (ICCV), 2017 IEEE International Conference
  on}, 2017.

\bibitem{zhu2017unpaired}
Jun-Yan Zhu, Taesung Park, Phillip Isola, and Alexei~A Efros.
\newblock Unpaired image-to-image translation using cycle-consistent
  adversarial networks.
\newblock In {\em Proceedings of the IEEE international conference on computer
  vision}, pages 2223--2232, 2017.

\end{thebibliography}
}

\end{document}